\documentclass[letterpaper, 10 pt, conference]{ieeeconf}
\IEEEoverridecommandlockouts
\usepackage{graphics} %

\usepackage{epsfig} %
\usepackage{mathptmx} %
\usepackage{times} %
\usepackage{amsmath} %
\usepackage{amssymb}  %
\usepackage{multirow}
\usepackage{color}
\usepackage{amsfonts}
\usepackage{xcolor}
\usepackage{tikz}
\usepackage{etex}
\usepackage{url}
\usepackage{footnote}
\usepackage{floatrow}
\usepackage{booktabs}
\usepackage{inputenc}
\usepackage{txfonts}
\usepackage{multicol}
\usepackage{siunitx}
\usepackage{rotating}
\usepackage{arydshln}
\usepackage{algorithm}
\usepackage{algorithmic}
\usepackage{subcaption}
\usepackage[noadjust]{cite}

\title{\LARGE \bf
Learning-aided Bigraph Matching Approach to Multi-Crew Restoration of Damaged Power Networks Coupled with Road Transportation Networks
}

\author{Nathan Maurer$^{1}$, Harshal Kaushik$^{2}$, Roshni Anna Jacob$^{2}$,  Jie Zhang$^{2}$, and Souma Chowdhury$^{1}$
\thanks{$^\dagger$ Corresponding Author, soumacho@buffalo.edu}
\thanks{$^{1}$ University at Buffalo, Buffalo, NY}
\thanks{$^{2}$ University of Texas at Dallas, Richardson, TX}%
\thanks{*This work was supported by the Office of Naval Research (ONR) grant N00014-21-1-2530. Any opinions, findings, conclusions, or recommendations expressed in this paper are those of the authors and do not necessarily reflect the views of ONR.}
\thanks{Copyright \textcopyright 2025 ASME. Personal use of this material is permitted. Permission from ASME must be obtained for all other uses, in any current or future media, including reprinting/republishing this material for advertising or promotional purposes, creating new collective works, for resale or redistribution to servers or lists, or reuse of any copyrighted component of this work in other works}
}

\begin{document}

\maketitle
\thispagestyle{empty}
\pagestyle{empty}

\begin{abstract}
The resilience of critical infrastructure networks (CINs) after disruptions, such as those caused by natural hazards, depends on both the speed of restoration and the extent to which operational functionality can be regained. Allocating resources for restoration, a combinatorial optimal planning problem that involves determining which crews will repair specific network nodes and in what order, is complicated by several factors: the connectivity between the CIN and the road transportation network used for travel by repair crews, the enormous scale and nonlinear behavior of these networks, and the uncertainty in repair times. This paper presents a novel graph-based formulation that merges two interconnected graphs, representing crew and transportation nodes and power grid nodes, into a single heterogeneous graph. To enable efficient planning, graph reinforcement learning (GRL) is integrated with bigraph matching. GRL is utilized to design the incentive function for assigning crews to repair tasks based on the graph abstracted state of the environment, ensuring generalization across damage scenarios. Two learning techniques are employed: a graph neural network trained using Proximal Policy Optimization and another trained via Neuroevolution. The learned incentive functions inform a bipartite graph that links crews to repair tasks, enabling weighted maximum matching for crew-to-task allocations. An efficient simulation environment that pre-computes optimal node-to-node path plans is used to train the proposed restoration planning methods. An IEEE 8500-bus power distribution test network coupled with a 21 sq km transportation network is used as the case study, with scenarios varying in terms of numbers of damaged nodes, depots and crews. Results demonstrate the approach’s generalizability and scalability across scenarios, with learned policies providing 3-fold better performance than random policies, while also outperforming optimization-based solutions in both computation time (by several orders of magnitude) and power restored.
\end{abstract}
\section{Introduction}
The increasing frequency of extreme weather events due to climate change and the consequent power outages in distribution networks (DNs), necessitates measures to improve resilience. Following the identification, isolation, and management of faults, the DN is to be restored to its pre-disruption state by repairing the damaged components. This requires optimally allocating repair crews or resources while considering transportation and power network constraints to maximize power restoration and the speed of recovery.  

Traditionally utilities perform these repairs based on experience, prior knowledge of critical components, and the operational state of the network. Such approaches may no longer be suitable for \lq\lq smart-grids\rq\rq\ that employ intelligent control algorithms for outage management. These automated systems dynamically alter network operations and shift restoration priorities. Additionally, DNs are often spread over large geographic areas which increase the scale of decision making. The heterogeneity of damaged nodes, varying degrees of damage, and the diverse types of crews and equipment required introduce additional complexity. As a result, restoration coordination becomes a high-dimensional, stochastic problem requiring efficient decision-making. Thus, the task of resource allocation and crew routing constitutes an NP-hard combinatorial optimization problem, making conventional approaches inefficient at scale.

Recently, mathematical programming approaches~\cite{arif2019repair,arif2017power,lei2019resilient} have been developed to provide decision support for distribution system operators in solving this optimization problem. However, these methods are scenario-specific and face limitations in scalability, as they struggle with increasing DN size and crew availability, as well as generalizability across different DNs and damage conditions. Furthermore, with the transition to ``smart grids'', there is a growing need for time-sensitive and online-deployable models to support real-time restoration efforts.

Restoration and repair of the distribution network rely on accurate fault location, isolation, and situational awareness. In traditional grids, this process typically involves manual inspection and confirmation through the dispatch of field crews or drones, leading to delays. In contrast, the smart grid introduces the concept of self-healing, enabling faster and more autonomous restoration. Critical to this is the leveraging of real-time measurements from Supervisory Control and Data Acquisition (SCADA) systems and Phasor Measurement Units (PMUs)~\cite{liu2013healing}, so that the network can perform intelligent state and topology estimation. This capability allows outage management systems (OMS) to be triggered remotely, enabling immediate initiation of restoration procedures. To realize the full potential of a self-healing grid, the restoration process must also be fast-acting, adaptive, and online deployable. An automated restoration framework minimizes downtime, leaving only the physical repair time, thereby preventing prolonged outages and avoiding cold load pickup issues~\cite{pang2023dynamic}.

This paper presents a multi-crew graph theoretic abstraction of the resource allocation problem and solves it through a combination of learned heuristics and maximal weighted (graph) matching. The power distribution network and the underlying transportation network used by the crews to reach the damaged nodes (i.e. the tasks) from their depots are modeled as an interconnected graph. The resource allocation and crew routing thus reduce to a multi-crew task allocation problem, which is represented as a bipartite graph or bigraph \cite{GHASSEMI2022103905}. Building on prior work in multi-robot task allocation, the problem is further formulated as a maximal weighted matching of a bipartite graph, where bipartite graph matching is applied to the task graph. The RL agent learns to assign weights to the edges in the bipartite graph to facilitate timely and efficient task and resource allocation.
\begin{figure*}[ht]
    \centering
    \includegraphics[width=0.99\textwidth]{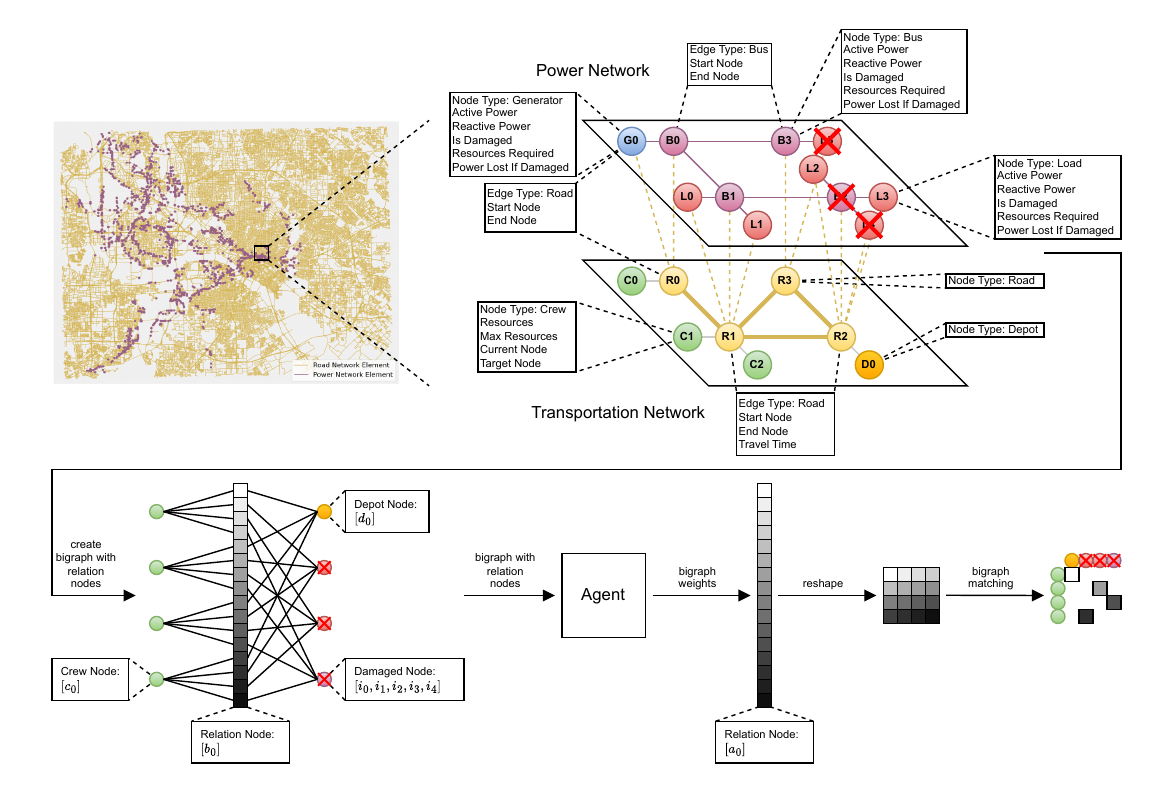}
    \caption{Information Flow Diagram For Learning Based Power Network Restoration}
    \label{fig:information_flow_diagram}
\end{figure*}

\subsection{Related Works}

The co-optimization of repairs and system restoration is a well known NP-hard combinatorial optimization problem. In~\cite{arif2017power}, a mixed-integer linear program was formulated considering the operation of the distribution network with crew routing. This approach was further extended to consider uncertainties using a stochastic mixed-integer linear program in~\cite{arif2018OptimizeServiceRestoration}. Authors in~\cite{arif2019repair} enhanced this methodology by incorporating unbalanced operation, modeling fault isolation constraints, and coordinating multiple (tree and line) crews in the distribution system repair and restoration problem. This work has been recently extended in~\cite{KPEC_HDK2024} by intertwining the transportation and power networks to solve combined resource allocation and capacitated vehicle routing problems. However, the optimization methods presented in these works struggle with scalability and lack generalizability to unseen scenarios, an expected limitation of traditional optimization techniques.

Graph Reinforcement Learning has been shown to be an effective method for solving multi-crew task allocation problems~\cite{Kool2019, Kaempfer2018LearningTM, khalil2017learning, Tolstaya2020MultiRobotCA, Paul_ICRA, paul2023efficient, krisshnakumar2023fast, kumar2023graph, paul2024learning} as well as power grid related planning problems \cite{jacob2024real}. These methods leverage graph structures to efficiently model and solve complex allocation tasks, particularly in dynamic and high-dimensional environments. Additionally, Bigraph Matching has been successfully applied to multi-crew task allocation, particularly in the context of Collective Transportation, as demonstrated in \cite{paul2024bigraph}.

Neuroevolution \cite{galvan2021neuroevolution,behjat2022adaptive} has also emerged as an effective approach for solving multi-crew task allocation problems \cite{krisshnakumarcomparative}.  Evolutionary methods, in general, have demonstrated promising performance in combinatorial optimization tasks~\cite{krisshnakumarcomparative, radhakrishnan2021evolutionary}, making them an alternative for addressing the complexities associated with large-scale task allocation and resource optimization.

\subsection{Key Contributions}
The overall objective of this paper is to present a learning approach for an incentive function that enables fast multi-crew task allocation via maximum weighted bigraph matching. This method is applied to the restoration of large-scale power grid networks with damaged nodes, while accounting for the transportation network used by repair crews to move between depots and task locations. The main contributions of this paper include: \textbf{1)} Abstracting the complex task and crew space that involves a multi-layered network (power and transportation networks) and crew state within it, as a single heterogeneous network. \textbf{2)} Developing a problem formulation and solution approach that poses the task/crew space as a bigraph, and performs allocation of tasks (which crew repairs which damaged power node) using weighted maximum matching. The weights, representing crew-to-task incentives, are automatically learned. \textbf{3)} Presenting two alternate learning approaches, graph neural networks trained by reinforcement learning (policy gradient techniques) and graph neural networks trained using neuroevolution. \textbf{4)} Demonstrating the proposed method for large-scale restoration planning in an 8,500-bus power grid coupled with a city-scale transportation network, and comparing its performance against random and heuristic baselines, as well as optimization based solutions. 
\section{Multi-Crew Task Allocation - Power Network Restoration}

The problem is modeled using a Reinforcement Learning environment that uses the Gymnasium Interface \cite{kwiatkowski2024gymnasium}.  A single heterogeneous graph represents the entirety of the environment and has three major subgraphs: the power network, the transportation network, and the network coupling.  The network coupling provides a location mapping between the transportation network and the power network.  The environment simulates power network disruption, power network repair, crew movement, resource logistics, and task allocation.

\subsection{Problem Formulation}
\label{sec:problem-formulation}

\subsubsection{Observations}
The observation space is an unweighted, undirected graph containing:
\begin{itemize}
    \item $R$ - set of relation nodes
    \item $C$ - set of crew nodes
    \item $D$ - set of depot nodes
    \item $I$ - set of initially damaged nodes
    \item $E_{R,C}$ - set of edges connecting relation nodes and crew nodes
    \item $E_{R,D}$ - set of edges connecting relation nodes and depot nodes
    \item $E_{R,I}$ - set of edges connecting relation nodes and initially damaged nodes
\end{itemize}

The observation graph is a bipartite graph with relation nodes inserted along each of it's edges.  The relation nodes are a convenience in implementation.
Relation nodes provide the following information to the agent:
\begin{itemize}
    \item $b_{0}$ (float) - travel time required for the Crew to reach the Target
\end{itemize}
Crew nodes provide the following information to the agent:
\begin{itemize}
    \item $c_{0}$ (float) - the number of resources the crew currently has
\end{itemize}
Initially damaged nodes provide the following information to the agent:
\begin{itemize}
    \item $i_{0}$ (boolean) - if the node is damaged
    \item $i_{1}$ (float) - remaining resources required for repair
    \item $i_{2}$ (float) - remaining time required for repair
    \item $i_{3}$ (float) - power lost if this is the only damaged node
    \item $i_{4}$ (boolean) - if the damaged node is connected to a powered node
\end{itemize}
Depot nodes provide the following information to the agent:
\begin{itemize}
    \item $d_{0}$ (float) - 0 (Placeholder value)
\end{itemize}
See Figure \ref{fig:information_flow_diagram} for a visualization of the observation space.

\subsubsection{Actions}
The values generated by a policy at the relation nodes are the actions.  The number of relation nodes, $n_r$ is computed by $n_r = n_c \times n_t$, where $n_c$ is the number of crews and $n_t$ is the number of targets.  See Figure \ref{fig:information_flow_diagram} for a visualization of the action space.

The relation node outputs are:
\begin{itemize}
    \item $a_{0}$ (float) - a bipartite graph weight
\end{itemize}

The output at a relation node is a single scalar, which represents the respective weight in the bipartite graph matrix.  The respective weight is determined using the corresponding crew row and target column.

\subsubsection{Reward} The reward $r$ is computed using the following equation: \\
\begin{equation*}
r = \frac{p - p_{\text{init}}}{p_{\text{max}} - p_{\text{init}}} \times \frac{t_{\text{step}}}{t_{\text{episode}}}
\end{equation*}
where:
\begin{itemize}
    \item $p$ is the power supplied to the network at the start of the step
    \item $p_{\text{init}}$ is the power supplied to the network at the start of the episode
    \item $p_{\text{max}}$ is the power supplied to the power network when the power network is undamaged
    \item $t_{\text{step}}$ is the duration of a timestep
    \item $t_{\text{episode}}$ is the duration of an episode
\end{itemize}
The cumulative reward for an episode is calculated as the ratio of the actual kWh supplied during the episode to the kWh that would have been supplied if the network had remained undamaged. This formulation ensures that the total cumulative reward for an episode remains between 0 and 1, which helps a user gauge how well a learning algorithm is performing. Maximizing this reward effectively corresponds to maximizing the power restored over time.

\subsection{Power Network Modeling}
The power distribution network is modeled using the open-source distribution system simulator (OpenDSS)~\cite{dugan2016reference}, a widely used tool for distribution network analysis. The power flow analysis and the resulting network state are also evaluated using OpenDSS. For circuit modifications, network element control, and other I/O operations, OpenDSSDirect~\cite{krishnamurthy2017opendssdirect}, a Python-based API is utilized. The specific power network model is loaded and simulated using the API, with the line outages simulated by opening/closing specific lines in the network. Power flow is evaluated both during outages and after repairs, where restoration is simulated by closing the affected lines. The key performance indicator in this study is the power served to the network, measured as the total power supplied to loads.

\subsection{Transportation Network Modeling}

The transportation network is modeled using a weighted directed graph with multiple Edges.  Nodes represent the points of interest along the road network. Edge weights represent the travel time for a vehicle moving at the speed limit from the origin to the destination of the directed edge. A crew node traverses the graph by moving along the shortest path to their target node.

\subsection{Power and Transportation Network Coupling}
Using OpenDSS, we simulated an 8,500-feeder network and obtained the local coordinates of all power network components. Next, we selected the primary level of the DN and mapped it onto a transportation network. This mapping process involves converting local coordinates from the primary power network into geodesic coordinates. The conversion is performed by applying specific x and y offsets relative to the bottom-left corner of the transportation network. The transportation network data is sourced from OpenStreetMap.

\subsection{Modeling Power Network Disruption: Damage and Repair Times}

Nodes in the power network are damaged by randomly selecting $n_{\text{damaged}}$ power nodes, where $n_{\text{damaged}}$ is a parameter of the environment.  A damaged nodes repair time, $t_{\text{repair}}$, is determined using a lognormal distribution with $\mu = -0.3072$ and $\sigma = 1.8404$.  $t_{\text{repair}}$ is truncated with a range of $[1,8]$.  Note that $t_{\text{repair}}$ is distinct from $p_{2}$ from Section \ref{sec:problem-formulation} in that $t_{\text{repair}}$ is the initial time required for repair whereas $p_{2}$ is the remaining time required for repair.

\subsection{Modeling Resource Logistics}

Each crew is initialized with 5 resources ($c_{1} = 5$).  If a crew is assigned to a depot node, it travels to that depot node.  Once the crew arrives at the depot node, the crews resources are immediately replenished.  If a crew has resources and is assigned to a damaged node, it navigates to and then repairs the damaged node.  The first step in repair is to drop off resources at the damaged node.  The second step in repair is to perform repairs on the damaged node, which requires that the resources required are present at the damaged node and that the crew stays at the damaged node for the time required to repair the damaged node.  The second step is only started once a sufficient number of resources have been placed at the damaged node.

\subsection{Modeling the Environment as a Heterogeneous Graph}
The environment graph is represented as one weighted directed graph with multiple edges.  The environment graph contains four subgraphs: the power network, the transportation network, the interconnect network, and the bipartite graph.

The power network is a subgraph of the environment graph and is equivalent to an unweighted undirected graph. The power network contains power nodes ( generator buses, consumer buses), and edges (distribution lines).  The power bus nodes are similar to the road nodes for the transportation network in that they are nodes of interest in the power network.

The transportation network is a subgraph of the environment graph and is a weighted directed graph with multiple edges.  The transportation network is comprised of road nodes, depot nodes, road-road edges, and road-depot edges.  The transportation network is traversable by crews.

The interconnect network graph is equivalent to an unweighted undirected graph.  It connects the power network to the transportation network and contains power nodes, road nodes, and power-road edges.  It is similar to the transportation network in that it is traversable by crews.

The bipartite graph is equivalent to an unweighted undirected graph.  It is composed of relation nodes, crew nodes, depot nodes, initially damaged nodes, relation-crew edges, relation-depot edges, and relation-damaged edges. The bipartite graph is comprised of two sets which are fully connected to one another.  The first set is composed of crew nodes, and the second set is composed of depot nodes and initially damaged nodes.  Relation nodes are inserted in each edge of this bipartite graph as a convenience of implementation.  The bipartite graph (relation nodes included) forms the observation space of the environment while the relation nodes alone form the action space.

\subsection{Modeling Task Allocation}
Actions are weights of a bipartite graph.  The bipartite graph is between crews and targets.  The two types of targets are damaged power nodes and depots.  Just before bipartite graph matching is performed, bipartite graph pre-processing is performed.  Bipartite graph pre-processing does the following:
\begin{itemize}
\item prevents crews from being assigned to damaged nodes that already have a crew assigned to them.
\item prevents crews from being assigned to targets that have already been repaired.
\item prevents crews with no resources from navigating to damaged nodes.
\item prevents crews with the maximum resources from navigating to depot nodes.
\end{itemize}

\subsection{Environment Stochasticity}
Each simulation step represents 1 hour.  The time available to each crew per timestep is computed by sampling from the normal random distribution, $\mathcal{N}(1, 0.1)$, and multiplying by 1 hour.  The implication is that a given crew will complete 1 hour worth of traveling and/or repairing, but may do more or less than that depending on the value sampled from the normal random distribution.
\begin{figure}[htbp!]
    \centering
    \includegraphics[width=0.95\textwidth]{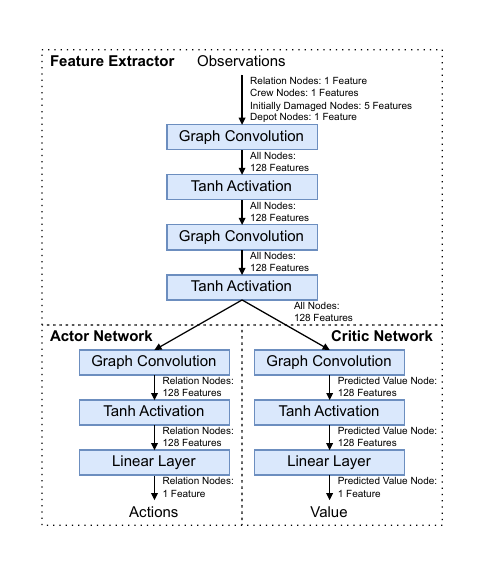}
    \caption{Graph Neural Network Policy architecture trained by Proximal Policy Optimization}
    \label{fig:actor_critic_feature_extractor_diagram}
\end{figure}

\begin{figure}[htbp!]
    \centering
    \includegraphics[width=0.95\textwidth]{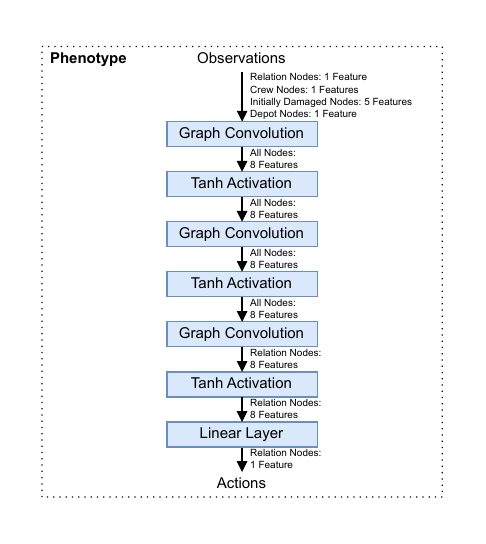}
    \caption{Neural Network Policy Architecture trained with Neuro-evolution approach}
    \label{fig:neuroevolution-perceptron}
\end{figure}
\section{Learning-aided Bigraph Matching Approach}

\subsection{Crew-Task Allocation via Bigraph Matching}

The Power Network Restoration Environment utilizes a bipartite graph in the assignment of crews to specific tasks. A bipartite graph (bigraph) is a graph with two distinct and non-overlapping vertex sets such that every edge in the graph connects a vertex from one set to a vertex in the other set. A weighted bigraph is a bigraph where the edges are weighted. An advantage of using bipartite graph matching for myopic task allocation is that each crew is assigned at most to one task and vice versa, hence automatically taking care of conflict resolution constraints in crew/repair task allocation. Weights in such a bigraph represents the value or incentive associated with pairing of nodes (vertices) across the two sets, which in our case represents the relative incentive for a crew (in the crew set) to be assigned the task (in the task set) that it is connected to.

Once a crew has been assigned to a task through this bipartite graph, they must complete it before being reassigned to another task. The purpose behind this is to avoid any potential disruptions or delays caused by frequent crew reassignments. This method promotes better organization, productivity, and overall efficiency of the crew assignments.

We construct a bigraph $\mathcal{G} = (V, E, \Omega)$, where the two types of nodes in $V$ are from the set of crews, $V^{C}$, and the set of tasks, $V^{T}$. $E$ represents the edges that connect each node in $V^{C}$ to each node in $V^{T}$. $\Omega$ is the weight matrix where $\Omega_{i,j}$ is the weight of the edge that connects $V^{C}_i$ to $V^{T}_j$.

Weighted bigraph matching is the process of selecting weights from the bigraph such that the sum of the weights is maximized subject to a couple constraints. The matching must be such that each node in one set is assigned at most to one node in the other set and vice versa. Popular methods for weight matching for bigraphs include the Hungarian Algorithm \cite{kuhn1955hungarian} and Karp Algorithm \cite{hopcroft1973n}.

\subsection{Incentive Learning via Graph Neural Networks}

In the Power Network Restoration Environment, incentive learning involves an agent learning how to assign values to each edge in the Bipartite Graph between the Crew and Target Nodes. In our implementation, each edge in the Bipartite Graph is represented as an Relation Node. The output embedding of each Relation Node is a single scalar value (-$\infty$, $\infty$). The graph neural network policy is used to compute the incentive values given the state of the operation at any point in time. 

Given the expected non-linearity of the reward function, large state and action spaces for incentive model (policy), and potential for overfitting, we explore two alternative ways for designing the GNN based incentive model. This includes a standard policy gradient reinforcement learning (RL) approach that is suited for efficient training of large policy models (providing adequate expressivity) and a neuroevolution approach that is suited to perform global search over non-convex reward functions even with parsimonious policy models. 

From the PPO approach, the Graph Neural Network (GNN) Feature Extractor, Actor Network, and Critic Network are composed of Graph Convolution Layers, Tanh activation functions, and a Linear Layer. The Linear Layer in the Actor Network is applied to each Relation node. The Actor Network's outputs are the weights of the bipartite graph. The critic network introduces a single Predicted Value Node.  All other nodes have a directed edge to the Predicted Value Node. The Linear Layer in the Critic Network is applied to the Predicted Value Node. The Critic Network's output is the output embedding of the Predicted Value Node which is the predicted value of a state-action pair.

The GNN Architecture used in the Neuroevolution approach is similar in type to the Neural Network Architecture used in the PPO approach, but varies in overall size and structure.  The Neuroevolution Architecture differs in that it does not include a Critic Network and reduces the number of features per node in the hidden layers.

For the learning based approaches, we use the Torch Geometric Graph Convolution Module \cite{Fey/Lenssen/2019, morris2019weisfeiler} shown below:

$\textbf{x}_{i}' = \textbf{W}_{1}\textbf{x}_{i} + \textbf{W}_{2}\sum_{j \in \mathcal{N}(i)} e_{j,i} \cdot \textbf{x}_{j}$
\\
where:
\begin{itemize}
    \item $i$ is the target node
    \item $j$ is the source node
    \item $e_{j,i}$ is the edge weight from source node $j$ to target node $i$
    \item $\textbf{x}_{i}$ is the target node embedding input
    \item $\textbf{x}_{j}$ is the source node embedding input
    \item $\textbf{x}_{i}'$ is the target node embedding output
    \item $\mathcal{N}(i)$ are the set of nodes that neighbor source node $i$
    \item $\textbf{W}_1$, $\textbf{W}_2$ are learned weight matrices
\end{itemize}

\begin{figure}[htbp!]
    \centering
    \begin{subfigure}[b]{0.95\textwidth}
        \centering
        \includegraphics[width=\textwidth]{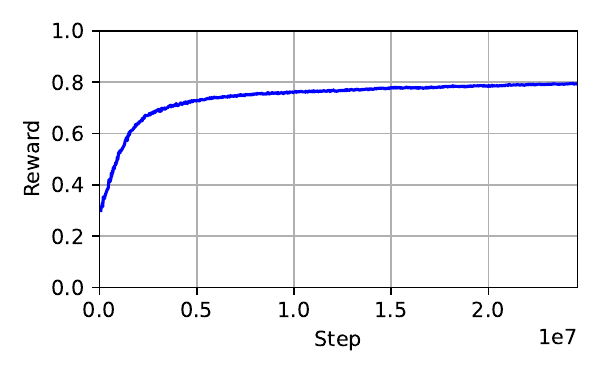}
        \caption{PPO Convergence Plot}
        \label{fig:convergence_plot_ppo}
    \end{subfigure}
    \begin{subfigure}[b]{0.95\textwidth}
        \centering
        \includegraphics[width=\textwidth]{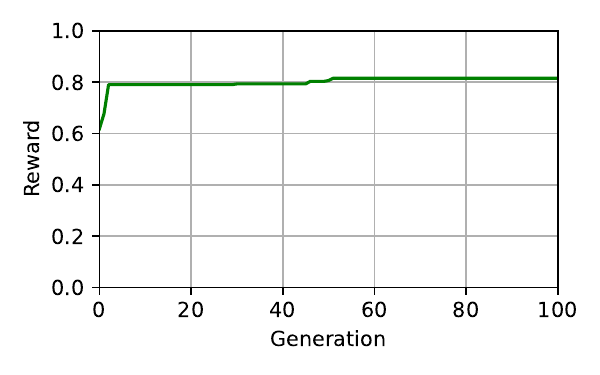}
        \caption{Neuro-evolution Convergence Plot}
        \label{fig:convergence_plot_neuroevolution}
    \end{subfigure}
    \caption{Training Convergence Plots}
    \label{fig:convergence_plots}
\end{figure}
\begin{figure*}[htbp!]
    \centering
    \begin{subfigure}[b]{0.45\textwidth}
        \centering
        \includegraphics[width=\textwidth]{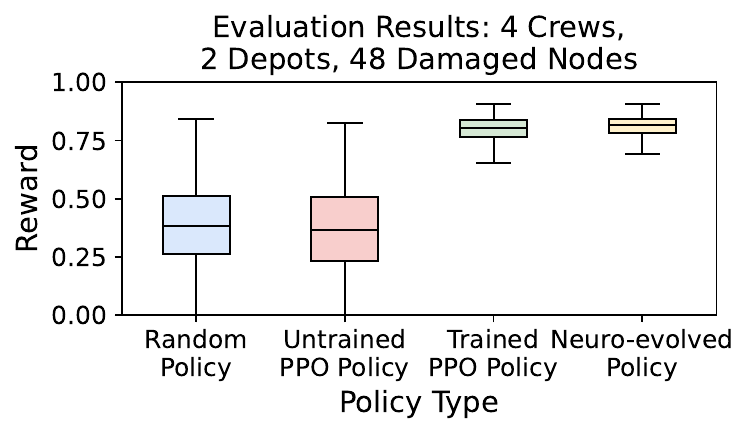}
        \caption{Unseen Scenarios for Evaluation A Environments}
        \label{fig:eval-plot-data-eval-4-2-48}
    \end{subfigure}
    \begin{subfigure}[b]{0.45\textwidth}
        \centering
        \includegraphics[width=\textwidth]{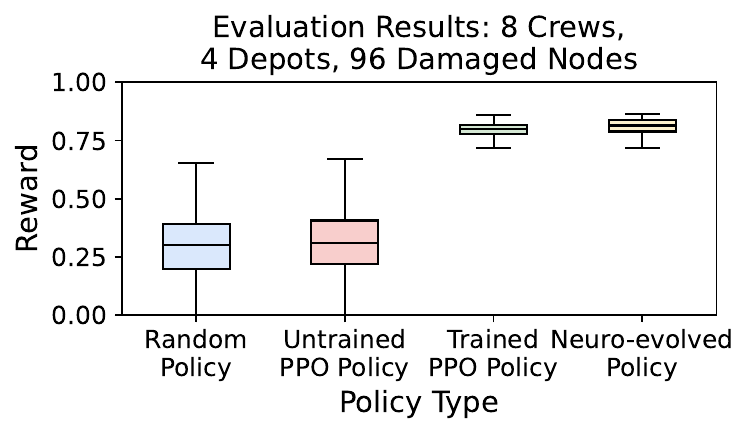}
        \caption{Unseen Scenarios for Evaluation B Environments}
        \label{fig:eval-plot-data-eval-8-4-96}
    \end{subfigure}
    \begin{subfigure}[b]{0.45\textwidth}
        \centering
        \includegraphics[width=\textwidth]{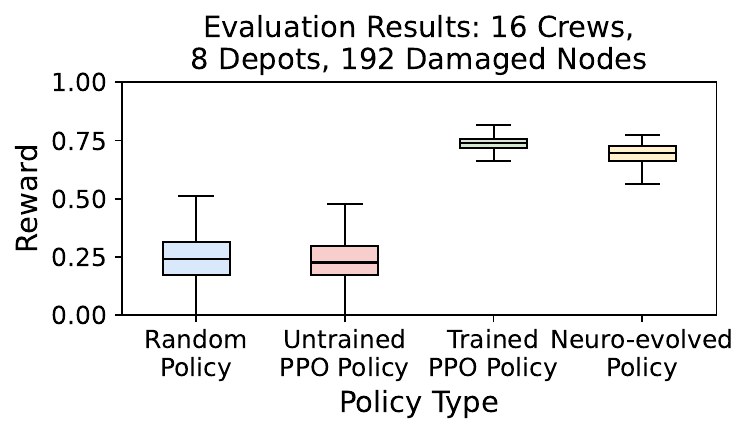}
        \caption{Unseen Scenarios for Evaluation C Environments}
        \label{fig:eval-plot-data-eval-16-8-192}
    \end{subfigure}
    \begin{subfigure}[b]{0.45\textwidth}
        \centering
        \includegraphics[width=\textwidth]{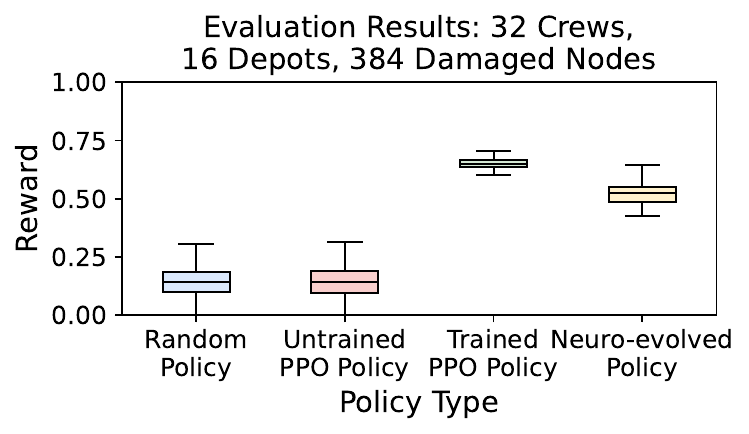}
        \caption{Unseen Scenarios for Evaluation D Environments}
        \label{fig:eval-plot-data-eval-32-16-384}
    \end{subfigure}
    \caption{Four different approaches evaluated on four unique sets of unseen scenarios}
    \label{fig:eval_plots}
\end{figure*}
\begin{figure}[htbp!]
    \begin{subfigure}[b]{0.95\textwidth}
        \centering
        \includegraphics[width=0.8\textwidth]{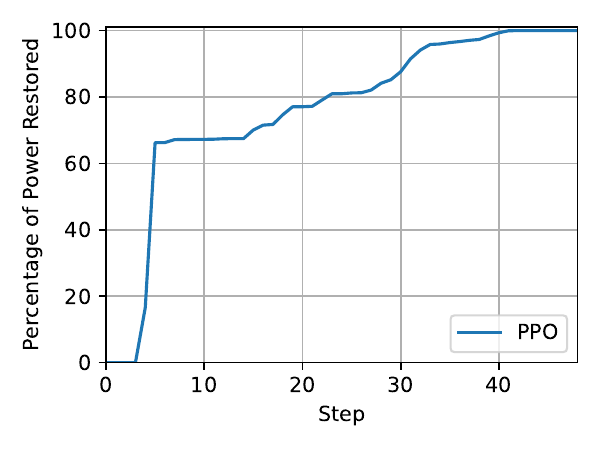}
        \caption{Sample Episode for Evaluation B - 8 crews, 4 depots, and 96 damaged nodes.}
    \end{subfigure}
    \begin{subfigure}[b]{0.95\textwidth}
        \centering
        \includegraphics[width=0.8\textwidth]{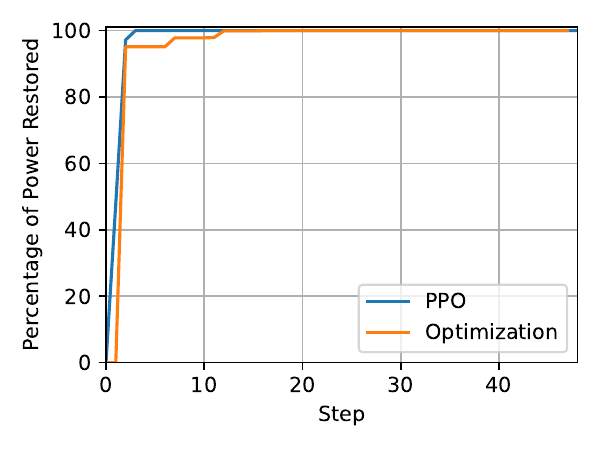}
        \caption{Sample Episodes for Evaluation OA - 2 crews, 3 depots, and 17 damaged nodes.}
    \end{subfigure}
    \begin{subfigure}[b]{0.95\textwidth}
        \centering
        \includegraphics[width=0.8\textwidth]{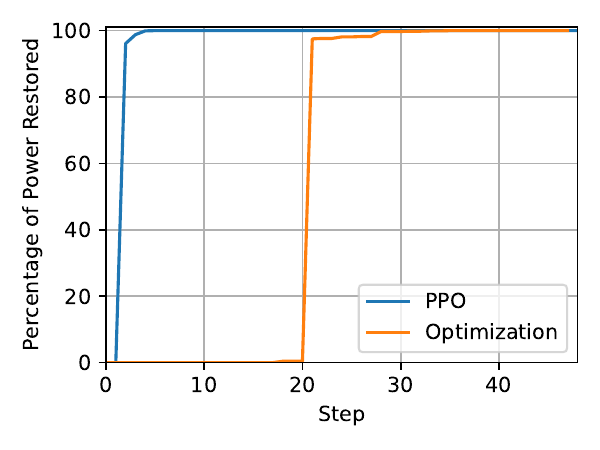}
        \caption{Sample Episodes for Evaluation OB - 2 crews, 3 depots, and 17 damaged nodes.}
    \end{subfigure}
    \caption{Percentage of Power Restored vs. Simulation Step for sample episodes.  Each Simulation Step is equivalent to 1 hour.}
    \label{fig:sample-episode-percentage-power-restored-plot}
\end{figure}
\section{Results}

\subsection{Evaluation Environment}

\subsubsection{OpenStreetMap Transportation Network}

For the transportation network, we utilize  a road network within the Dallas-Forth Worth (DFW) area. Using Networkx  and OpenSteetMap \cite{boing_Osmnx2017} packages, we obtain the real road network information in the DFW area. We chose this area based on the spatial requirements of an 8500-node test feeder network, which requires a minimum of 2100 square kilometers. Accordingly, we selected a corresponding region within the DFW area. From NetworkX and OpenStreetMap, we obtain detailed information about graph nodes and edges, including attributes such as OSMIDs, latitude, longitude, x, and y coordinates of nodes, graph connectivity, edge lengths, road types (one-way/two-way), maximum speeds, and the number of lanes. For our purposes, we select only the node coordinates (latitude and longitude) and the connectivity of edges within the graph.

\subsubsection{IEEE 8500 Power Network}

A benchmark distribution test network, specifically the IEEE 8500-node test feeder, is employed to validate the proposed model~\cite{IEEE8500feeder}. This test feeder is a large radial distribution network that closely resembles a real-world North American feeder, making it effective to demonstrate the practicality of the restoration model. The feeder includes both primary (medium voltage) and secondary (low voltage) levels. Given the critical nature of primary line failures, the network has been reduced down to its medium voltage primary level by aggregating the loads at the distribution transformers. The network is further evaluated using the OpenDSSDirect API, with its state represented as a NetworkX graph.

\subsubsection{Coupling of Transportation Network with Power Network}

We have the local coordinates available for the power network components from OpenDSS. Mapping the local nodes from the power network to the transportation network is straightforward. We converted the local coordinates (from the power network) into geodesic coordinates by adjusting them with specific x and y offsets, based on the bottom-left corner of the transportation network. Note that this information is available to us from OpenStreetMap.
\subsection{Planning Methods}

The solutions examined are a GNN based Policy learned via PPO, a GNN based Policy learned via Neuro-Evolution, a Random Action Policy, and Optimization Solutions from mixed integer programming. The Random Action Policy and Optimization Solutions serve as baseline methods that the PPO Policy and Neuro-evolved Policy are compared against.  Additionally, an untrained PPO Policy is included to highlight the performance gains achieved through training and to serve as another baseline.  The PPO Policy is selected because it is trained using a policy gradient method, while the Neuro-evolved Policy is selected because it is trained using a gradient-free learning method.

\subsubsection{PPO Policy}

Figure \ref{fig:actor_critic_feature_extractor_diagram} shows the network architecture for the PPO Policy.  The Untrained PPO Policy is the PPO Policy Network before it is trained.  This method serves as a lower bound on the performance expected from the other methods.  The Trained PPO Policy is the PPO Policy Network after it is trained.  Table \ref{tab:learning_hyperparameters} shows the PPO training hyperparameters.  The Stable Baselines 3 \cite{stable-baselines3} implementation of PPO is used for training.  Figure \ref{fig:actor_critic_feature_extractor_diagram} shows the network architecture.

By default, the Stable Baselines 3 implementation of Proximal Policy Optimization outputs a different standard deviation for each policy output when the action space is continuous.  This is modified such that the actions are sampled from a gaussian distribution where each action uses the same learned standard deviation.  The issue this solves is that when the action space size changes, our policy does not need to learn new standard deviation values.

\subsubsection{Neuro-evolved Policy}
The Neuro-evolution approach uses a neural network architecture that is similar to the neural network architecture used by the Proximal Policy Optimization Feature Extractor and Actor.  The key differences between the two are that the Neuroevolution architecture does not use the value network and uses less channels than the PPO network architecture.  Figure \ref{fig:neuroevolution-perceptron} displays a visual representation of the Neuroevolution Network Architecture.

\subsubsection{Random Action Policy}

The Random Action Policy determines actions by sampling from the action space.  This method serves as a lower bound on the performance expected from the other methods.

\subsubsection{Optimization}
The optimization problem is formulated as a mixed-integer programming (MIP) model, where the primary objective is to assign repair crews to damaged nodes based on their power restoration potential (measured in kW) and estimated repair times (in hour). The second objective focuses on determining the most efficient travel routes for each crew by solving a Vehicle Routing Problem (VRP). The key constraints include ensuring that each crew starts from a designated depot and returns to either the same or another depot upon completing its assigned tasks. Additionally, each crew is equipped with a repair kit of finite capacity, which must be sufficient to meet the repair demands at the damaged nodes. When the repair capacity is depleted, the crew must return to the depot for replenishment before continuing restoration efforts.

We used Gurobi Optimizer version 11.0.0 on a machine with an Intel Core i7-1365U CPU. The Gurobi solver employed a combination of branch-and-cut and simplex methods to solve the optimization problem.

Due to the large-scale nature of the underlying transportation and power network, solving this optimization problem to optimality with Gurobi resulted in significant scalability issues. These issues caused longer computation times as the problem size increased, particularly with the growing number of damaged nodes.
\begin{table}[htbp!]
\centering
\caption{Environment Configurations}
\begin{tabular}{|l|c|c|c|}
\hline
\textbf{Configuration} & \textbf{\# Crews} & \textbf{\# Depots} & \textbf{\# Damaged Nodes} \\ \hline
Train & 8 & 4 & 96 \\ \hline
Eval A & 4 & 2 & 48 \\ \hline
Eval B & 8 & 4 & 96 \\ \hline
Eval C & 16 & 8 & 192 \\ \hline
Eval D & 32 & 16 & 384 \\ \hline
Eval OA & 2 & 3 & 5 \\ \hline
Eval OB & 2 & 3 & 17 \\ \hline
\end{tabular}
\label{tab:environment}
\end{table}
\subsection{Training}
\subsubsection{Training Details}

Training used of 32 unique environments.  Each episode represents a 48 hour duration where each step in the episode is equivalent to 1 hour. All training environments have 8 crews, 4 depots, 96 damaged nodes, and a 48 hour time limit.  For each of the 32 unique environments, the damaged nodes were randomly sampled from the Power Network nodes and the depots were placed next to random nodes sampled from the Transportation Network nodes.

\subsubsection{Training Performance}

From Figure \ref{fig:convergence_plot_ppo}, PPO improves throughout the training process, although the rate of improvement decreases with time.  Each data point is the average score of the policy being trained on 256 episodes (32 unique environments, 8 episodes per unique environment). The number of training iterations was set at 2,000 (512,000 episodes, 24,576,000 timesteps). Table \ref{tab:learning_hyperparameters} contains PPO hyperparameters that were modified from their defaults in the Stable Baselines 3 implementation. Training took 48.65 hours on a system with an AMD 5950X CPU and an Nvidia 3090 GPU.

From Figure \ref{fig:convergence_plot_neuroevolution}, Neuro-evolution first reaches the reward at which it will converge at generation 51.  The training remains at that level until the end at generation 100.  Each unique environment was run 8 times per genome evaluation for a total of 256 episodes per iteration (12,288 timesteps per iteration).  Each datapoint from Figure \ref{fig:convergence_plot_neuroevolution} is the average reward obtained by the best genome over 256 episodes (32 unique environments, 8 episodes per unique environment).  The number of genomes evaluated during the Neuroevolution training process is 3450, which equates to 883,200 episodes or 42,393,600 timesteps.  Training took 32.19 hours on a system with an AMD 5950X CPU and an Nvidia 3090 GPU.
\begin{table}[htbp!]
\centering
\caption{Learning Algorithm Hyperparameters}
\begin{tabular}{|l|l|}
\hline
\multicolumn{2}{|c|}{\textbf{PPO Hyperparameters}} \\ \hline
\textbf{Parameter} & \textbf{Value} \\ \hline
Learning Rate & 1e-5 \\ \hline
Batch Size & 384 \\ \hline
\# Steps per Iteration & 12,288 \\ \hline
Learning Iterations & 2000 \\ \hline
\multicolumn{2}{|c|}{\textbf{NE Hyperparameters}} \\ \hline
\textbf{Parameter} & \textbf{Value} \\ \hline
\# Generations & 100 \\ \hline
Population Size & 50 \\ \hline
Elite Ratio & 0.01 \\ \hline
Crossover Probability & 0.5 \\ \hline
Mutation Probability & 0.1 \\ \hline
\end{tabular}
\label{tab:learning_hyperparameters}
\end{table}
\subsection{Testing Performance and Comparisons}
From Figure \ref{fig:eval_plots}, as the number of crews and number of damaged nodes increases, the performance of all of the methods decreases.  However, it is notable that the trained PPO Policy and the Neuro-evolved Policy both perform significantly better than the random and untrained policy in all scenarios.  Both the trained PPO Policy and the trained Neuro-evolved Policy both scale well, indicating that a policies learned on smaller scenarios can be applied to larger scenarios.  All trained policies were trained on the 32 training scenarios that had 8 crews, 4 depots, and 96 damaged nodes.  The only policy changes required are resizing the observation space and action space.

From Table \ref{tab:optimization_comparison}, the forward inference for the PPO Policy and Neuro-evolved Policy is significantly faster than the time taken by optimization to solve the same problem.  From Table \ref{tab:optimization_comparison}, the PPO Policy and Neuro-evolved Policy performs better on Environment OA and both perform significantly better on Environment OB.  We can see that as the problem size increases, the performance decreases and the computation time increase drastically for the optimization.

Figure \ref{fig:sample-episode-percentage-power-restored-plot} shows the total power restored to the power network over the course of sample episodes.  From the plots, it is clear that the learned PPO Controller outperforms the Optimization Controller in terms of power restored.
\begin{table}[htbp!]
\centering
\caption{Optimization and Learned Policy Compute Time Comparisons}
\begin{tabular}{|l|l|l|l|}
\hline
\multicolumn{4}{|c|}{\textbf{Episode Reward}} \\ \hline
\textbf{Env Type} & \textbf{Optimization} & \textbf{PPO Policy} & \textbf{NE Policy} \\ \hline
Eval OA & 0.9510 & 0.9889 & 0.9888 \\ \hline
Eval OB & 0.5595 & 0.9773 & 0.9770 \\ \hline
\multicolumn{4}{|c|}{\textbf{Compute Time}} \\ \hline\textbf{Env Type} & \textbf{Optimization} & \textbf{PPO Policy} & \textbf{NE Policy} \\ \hline
Eval OA & 97.0000 s & 0.0165 s & 0.0080 s \\ \hline
Eval OB & 1852.8800 s & 0.0171 s & 0.0080 s \\ \hline
\multicolumn{4}{|c|}{\textbf{System Specs}} \\ \hline\textbf{Comp. Type} & \textbf{Optimization} & \textbf{PPO Policy} & \textbf{NE Policy} \\ \hline
CPU & i7-1365U & 5950X & 5950X \\ \hline
GPU & NA & 3090 & 3090 \\ \hline

\end{tabular}
\label{tab:optimization_comparison}
\end{table}
\section{Conclusion}


This paper presented a novel graph-based formulation of the crew allocation problem for repairing damaged nodes (e.g., caused by natural hazards) in the power distribution network to be restored, while also considering the road transportation network used by crews to move between depots and repair task locations. The solution approach to this inherently combinatorial optimal planning problem then integrates graph neural networks, trained by PPO and neuroevolution methods (as alternatives), to design incentive functions that reflect the suitability of allocating a specific repair task to a specific crew at any given point of time over a 48 hour repair operation period. This incentive functions are used by a provably optimal and interpretable bigraph matching approach to allocate tasks in a manner that generalizes across various damage scenarios. The simulation environment used to train and test this method accounted for travel times, stochastic repair times, repair resources, and power delivered to customers pre/post repair.

Both approaches to learning the GNN based policies for incentive computations were found to converge to similar reward values during training. Evaluation of this new restoration planning approach (integrating graph learning with bigraph matching) over unseen test scenarios of relatively unprecedented scale demonstrated both generalizability and scalability across varying numbers of crews, depots and damaged nodes. Compared to random crew-to-repair-task allocations, the performance of the proposed approach was found to be 3 times better on average in terms of the restored power metric used as the performance metric. The PPO trained policy was found to provide slightly better generalizability to larger scenarios compared to the neuroevolution trained policy, while the latter allowed a much more compact policy model to be used with success. In comparison to optimization (MIP) based solutions, which could feasibly be obtained only for the smaller scenarios, the proposed (graph learning and bigarph matching based) method was found to be orders of magnitude faster in execution while providing slightly or significantly better performance.

Future extension of the proposed method can explore additional complexities such as discrete heterogeneity of crews and repair tasks, disruptions to the transportation network, and partial observability of tasks. In its current form, the presented method works on a given power/transportation network couple, and further work is also required to analyze the additional computing cost required to transfer the learning architecture to other power/transportation network couples. These entensions, along with consideration of bidirectional coupling typical of CINs, e.g., where power disruptions also affect transportation disruptions, could provide a more comprehensive understanding of the potential of this novel graph-abstracted and learning aided approach to restoration planning. 

\bibliographystyle{IEEEtran}
\bibliography{asmeconf-sample}

\end{document}